\def\year{2021}\relax
\documentclass[letterpaper]{article} % DO NOT CHANGE THIS
\usepackage{aaai21}  % DO NOT CHANGE THIS
\usepackage{times}  % DO NOT CHANGE THIS
\usepackage{helvet} % DO NOT CHANGE THIS
\usepackage{courier}  % DO NOT CHANGE THIS
\usepackage[hyphens]{url}  % DO NOT CHANGE THIS
\usepackage{graphicx} % DO NOT CHANGE THIS
\usepackage{natbib}  % DO NOT CHANGE THIS AND DO NOT ADD ANY OPTIONS TO IT
\usepackage{caption} % DO NOT CHANGE THIS AND DO NOT ADD ANY OPTIONS TO IT
\usepackage{amsthm}
\usepackage{amsmath, amssymb}
\usepackage[utf8]{inputenc} % allow utf-8 input
\usepackage[T1]{fontenc}    % use 8-bit T1 fonts
\usepackage{url}            % simple URL typesetting
\usepackage{booktabs}       % professional-quality tables
\usepackage{amsfonts}       % blackboard math symbols
\usepackage{nicefrac}       % compact symbols for 1/2, etc.
\usepackage{lipsum}
\usepackage{mathtools}
\usepackage{subcaption}
\usepackage[switch]{lineno}
\DeclareMathOperator*{\argmax}{arg\,max}

\DeclarePairedDelimiter\ceil{\lceil}{\rceil}

\newtheorem{theorem}{Theorem}
\newtheorem{lemma}[theorem]{Lemma}

\newtheorem{definition}{Definition}
\newtheorem{corollary}[theorem]{Corollary}
\newtheorem{proposition}[theorem]{Proposition}

\newcounter{subdefinition}[theorem]
\renewcommand{\thesubdefinition}{\thedefinition.\arabic{subdefinition}}
\newenvironment{subdefinition}{
        \refstepcounter{subdefinition}
        \par\noindent
        \textbf{\upshape Definition \thesubdefinition}%
}{}

\usepackage{xpatch}
\makeatletter
\AtBeginDocument{\xpatchcmd{\@thm}{\thm@headpunct{.}}{\thm@headpunct{}}{}{}}
\makeatother

\graphicspath{ {./images/} }

\newcommand\numberthis{\addtocounter{equation}{1}\tag{\theequation}}

\title{`Less Than One'-Shot Learning:\\ Learning N Classes From M<N Samples}

\author{
    Ilia Sucholutsky, Matthias Schonlau
}
\affiliations{
    Department of Statistics and Actuarial Science\\
    University of Waterloo\\
    Waterloo, Ontario, Canada
    \textsuperscript{\rm 1}isucholu@uwaterloo.ca

}
\begin{document}
\nocopyright
%\linenumbers
\maketitle
\begin{abstract}
 Deep neural networks require large training sets but suffer from high computational cost and long training times. Training on much smaller training sets while maintaining nearly the same accuracy would be very beneficial. In the few-shot learning setting, a model must learn a new class given only a small number of samples from that class. One-shot learning is an extreme form of few-shot learning where the model must learn a new class from a single example. We propose the `less than one'-shot learning task where models must learn $N$ new classes given only $M<N$ examples and we show that this is achievable with the help of soft labels. We use a soft-label generalization of the k-Nearest Neighbors classifier to explore the intricate decision landscapes that can be created in the `less than one'-shot learning setting. We analyze these decision landscapes to derive theoretical lower bounds for separating $N$ classes using $M<N$ soft-label samples and investigate the robustness of the resulting systems.  %Our code and a live web demo can be found at \url{https://github.com/ilia10000/SLkNN}
\end{abstract}

% keywords can be removed

\section{Introduction}

\begin{figure}[t!]
    \centering
    \includegraphics[width=0.48\textwidth]{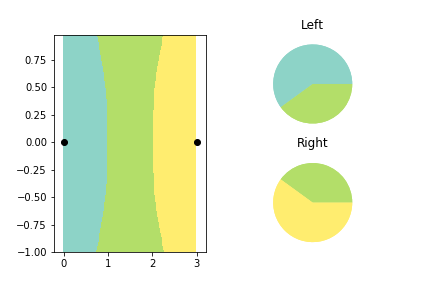}
    \caption{A SLaPkNN classifier is fitted on 2 soft-label prototypes and partitions the space into 3 classes. The soft label distribution of each prototype is illustrated by the pie charts.}
    \label{fig:2points_3classes}
\end{figure}

Deep supervised learning models are extremely data-hungry, generally requiring a very large number of samples to train on. Meanwhile, it appears that humans can quickly generalize from a tiny number of examples~\citep{lake2015human}. Getting machines to learn from `small' data is an important aspect of trying to bridge this gap in abilities. Few-shot learning (FSL) is one approach to making models more sample efficient. In this setting, models must learn to discern new classes given only a few examples per class~\citep{lake2015human, snell2017prototypical, wang2020generalizing}. Further progress in this area has enabled a more extreme form of FSL called one-shot learning (OSL); a difficult task where models must learn to discern a new class given only a single example of it~\citep{fei2006one, vinyals2016matching}. In this paper, we propose `less than one'-shot learning (LO-shot learning), a setting where a model must learn $N$ new classes given only $M<N$ examples, less than one example per class. At first glance, this appears to be an impossible task, but we both theoretically and empirically demonstrate feasibility. As an analogy, consider an alien zoologist who arrived on Earth and is being tasked with catching a unicorn. It has no familiarity with local fauna and there are no photos of unicorns, so humans show it a photo of a horse and a photo of a rhinoceros, and say that a unicorn is something in between. With just two examples, the alien has now learned to recognize three different animals. The type of information sharing that allows us to describe a unicorn by its common features with horses and rhinoceroses is the key to enabling LO-shot learning. In particular, this unintuitive setting relies on \textit{soft labels} to encode and decode more information from each example than is otherwise possible. An example can be seen in Figure~\ref{fig:2points_3classes} where two samples with soft labels are used to separate a space into three classes. 
There is already some existing evidence that LO-shot learning is feasible. \citet{sucholutsky2019soft} showed that it is possible to design a set of five soft-labelled synthetic images, sometimes called `prototypes', that train neural networks to over 90\% accuracy on the ten-class MNIST task. We discuss the algorithm used to achieve this result in more detail in the next section. In this paper, we aim to investigate the theoretical foundations of LO-shot learning so we choose to work with a simpler model that is easier to analyze than deep neural networks. Specifically, we propose a generalization of the k-Nearest Neighbors classifier that can be fitted on soft-label points.  We use it to explore the intricate decision landscapes that can still be created even with an extremely limited number of training samples. We first analyze these decision landscapes in a data-agnostic way to derive theoretical lower bounds for separating $N$ classes using $M<N$ soft-label samples. Unexpectedly, we find that our model fitted on just two prototypes with carefully designed soft labels can be used to divide the decision space into any finite number of classes. Additionally, we provide a method for analyzing the stability and robustness of the created decision landscapes. We find that by carefully tuning the hyper-parameters we can elicit certain desirable properties. We also perform a case study to confirm that soft labels can be used to represent training sets using fewer prototypes than there are classes, achieving large increases in sample-efficiency over regular (hard-label) prototypes. In extreme cases, using soft labels can even reduce the minimal number of prototypes required to perfectly separate $N$ classes from $\mathcal{O}(N^2)$ down to $\mathcal{O}(1)$.

The rest of this paper is split into three sections. In the first section we discuss related work. In the second section we explore decision landscapes that can be created in the `less than one'-shot learning setting and derive theoretical lower bounds for separating $N$ classes using $M<N$ soft-label samples. We also examine the robustness of the decision landscapes to perturbations of the prototypes. In the final section we discuss the implications of this paper and propose future directions for research.

\section{Related Work}

\citet{DD} showed that Dataset Distillation (DD) can use backpropagation to create small synthetic datasets that train neural networks to nearly the same accuracy as when training on the original datasets. Networks can reach over 90\% accuracy on MNIST after training on just one such distilled image per class (ten in total), an impressive example of one-shot learning. \citet{sucholutsky2019soft} showed that dataset sizes could be reduced even further by enhancing DD with soft, learnable labels.  Soft-Label Dataset Distillation (SLDD) can create a dataset of just five distilled images (less than one per class) that trains neural networks to over 90\% accuracy on MNIST. In other words, SLDD can create five samples that allow a neural network to separate ten classes. To the best of our knowledge, this is the first example of LO-shot learning and it motivates us to further explore this direction. 

However, we do not focus on the process of selecting or generating a small number of prototypes based on a large training dataset. There are already numerous methods making use of various tricks and heuristics to achieve impressive results in this area. These include active learning~\citep{active1, active2}, core-set selection~\citep{coreset3, coreset1, coreset2}, pruning~\citep{prune1}, and kNN prototype methods ~\citep{bezdek2001nearest,triguero2011taxonomy, garcia2012prototype, kusner2014stochastic} among many others. There are even methods that perform soft-label dataset condensation~\citep{ruta2006dynamic}. We do not explicitly use any of these methods to derive our soft-label prototypes. We instead focus on theoretically establishing the link between soft-label prototypes and `less than one'-shot learning. 

Our analysis is centered around a distance-weighted kNN variant that can make use of soft-label prototypes. Distance-weighted kNN rules have been studied extensively~\citep{dudani1976distance,macleod1987re,gou2012new, yigit2015abc} since inverse distance weighting was first proposed by~\citet{shepard1968two}. Much effort has also gone into providing finer-grained class probabilities by creating soft, fuzzy, and conditional variants of the kNN classifier~\citep{mitchell2001soft,el2006study,thiel2008classification,el2010semi, kanjanatarakul2018evidential, wang2019regularized, gweon2019k}. We chose our kNN variant, described in the next section, because it works well with soft-label prototypes but remains simple and easy to implement.

Several studies have been conducted on the robustness and stability of kNN classifiers. \citet{el2006study} found that that using soft labels with kNN resulted in classifiers that were more robust to noise in the training data. \citet{sun2016stabilized} proposed a nearest neighbor classifier with optimal stability based on their extensive study of the stability of hard-label kNN classifiers for binary classification. Our robustness and stability analyses focus specifically on the LO-shot learning setting which has never previously been explored.

\section{`Less Than One'-Shot Learning}
We derive and analyze several methods of configuring $M$ soft-label prototypes to divide a space into $N$ classes using the distance-weighted SLaPkNN classifier. All proofs for results in this section can be found in the appendix contained in the supplemental materials. 

% \begin{itemize}
%     \item kNN decision boundaries are locally linear, SLaPkNN ones are not
%     \item kNN approaches double bayes rate, what about in prototype case? slapknn case?
% \end{itemize}
\subsection{Definitions}
\begin{definition}
A \textbf{hard label} is a vector of length $N$ representing a point's membership to exactly one out of $N$ classes. 
\begin{align*}
y^{hard}={e}_i=\begin{bmatrix}
           0 \;
           ...  \;
           0 &
           1 &
           0 \;
           ... \;            
           0 
\end{bmatrix}^T
\end{align*}
\end{definition}
Hard labels can only be used when each point belongs to exactly one class.  If there are $n$ classes, and some point $x$ belongs to class $i$, then the hard label for this point is the $i^{th}$ unit vector from the standard basis. \\

\begin{definition}
A \textbf{soft label} is the vector-representation of a point's simultaneous membership to several classes. Soft labels can be used when each point is associated with a distribution of classes.  We denote soft labels by $y^{soft}$.\\ 
\begin{subdefinition}
A \textbf{probabilistic (soft) label} is a soft label whose elements form a valid probability distribution. \\
\begin{align*}
    \forall i \in \{1,...,n\}\; y^{soft}_i \geq 0\\
    \sum_{i=1}^n y^{soft}_i = 1
\end{align*}
\end{subdefinition}
\begin{subdefinition}
An \textbf{unrestricted (soft) label} is a soft label whose elements can take on any real value, including negative values.
\end{subdefinition}
\end{definition}
\begin{definition}
A \textbf{soft-label prototype (SLaP)} is a pair of vectors (X,Y) where X is a feature (or location) vector, and Y is the associated soft label.
\end{definition}

\begin{figure}[h!]
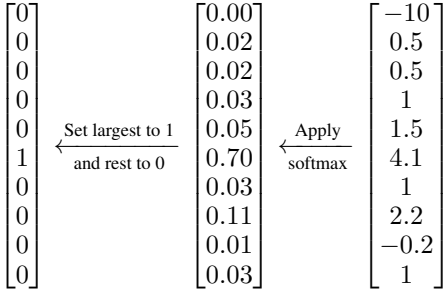

\begin{align*}
  \begin{bmatrix}
           0 \\
           0 \\
           0 \\
           0 \\
           0 \\
           1 \\
           0 \\
           0 \\
           0 \\
           0 \\
\end{bmatrix}
\xleftarrow[\text{and rest to 0}]{\text{Set largest to 1}}
\begin{bmatrix}
           0.00 \\
           0.02 \\
           0.02 \\
           0.03 \\
           0.05 \\
           0.70 \\
           0.03 \\
           0.11 \\
           0.01 \\
           0.03 \\
\end{bmatrix}
\xleftarrow[\text{softmax}]{\text{Apply}}
\begin{bmatrix}
           -10 \\
           0.5 \\
           0.5 \\
           1 \\
           1.5 \\
           4.1 \\           
           1 \\
           2.2 \\
           -0.2 \\
           1 \\
\end{bmatrix}
\end{align*}
\caption{\textbf{Left}: An example of a `hard' label where the sixth class is selected. \textbf{Center}: An example of a soft label restricted to being a valid probability distribution. The sixth class has the highest probability. \textbf{Right}: An example of an unrestricted soft label. The sixth class has the highest weight. `Hard' labels can be derived from unrestricted `soft' labels by applying the softmax function and then setting the highest probability element to 1, and the rest to 0.}
\label{soft_label_example}
\end{figure}

A probabilistic label can be derived from an unrestricted label by applying the softmax function. A hard label can be derived from a probabilistic label by applying the argmax function (setting the element with the highest associated probability to 1 and the remaining elements to 0). We illustrate this in Figure~\ref{soft_label_example}.

When using the classical k-Nearest Neighbors (kNN) classifier rule to partition a space, it is clear that at least $n$ points are required to separate $n$ classes (i.e. create $n$ partitions). kNN uses only hard labels, so each of these points, or \textbf{prototypes}, contains information only about the single class to which it is assigned. We investigate whether a soft-label prototype generalization of kNN can more efficiently separate $n$ classes by using only $m<n$ soft-label prototypes.
% \begin{definition}
% The \textbf{soft-label prototype k-Nearest Neighbors (SLaPkNN)} classification rule takes the sum of the label vectors of the $k$-nearest prototypes to target point $x$. $x$ is then assigned to the class corresponding to the largest value of the resulting vector. \\
% More formally, assume we are given a set of $M$ soft-label prototypes representing a dataset with $N$ classes. let $S = (X_1,Y_1), ..., (X_M, Y_M)$ be the prototypes available for training where $X_i$ is the position of the $i^{th}$ prototype and $Y_i$ is its soft label ($Y_i$ is a vector of length $N$). Let $x$ be the position of the target point that we wish to classify. Compute  $D=\{d(X_i,x)\}_{i=1,...,M}$, the set of distances between each prototype and the target. Let $S'=(X_{(1)},Y_{(1)}), ...,(X_{(M)}, Y_{(M)}) $ be a reordering of the prototypes such that $d(X_{(1)},x)\leq ... \leq d(X_{(M)},x)$. Then the sum of the $k$-nearest prototype labels is $Y^*=\sum_{i=1}^k Y_{(i)}$ and $x$ is assigned to class $C^{SLaPkNN}(x) = \argmax_j Y^*_j$ where $Y^*_j$ is the $j^{th}$ element of $Y^*$.

% \end{definition}

% \begin{remark}
% kNN is a special case of SLaPkNN where each prototype is assigned a hard label.
% \end{remark}
\begin{definition}
The \textbf{distance-weighted soft-label prototype k-Nearest Neighbors (SLaPkNN)} classification rule takes the sum of the label vectors of the $k$-nearest prototypes to target point $x$, with each prototype weighted inversely proportional to its distance from $x$. $x$ is then assigned to the class corresponding to the largest value of the resulting vector. \\
More formally, assume we are given a set of $M$ soft-label prototypes representing a dataset with $N$ classes. Let $S = (X_1,Y_1), ..., (X_M, Y_M)$ be the prototypes available for training where $X_i$ is the position of the $i^{th}$ prototype and $Y_i$, a vector of length $N$, is its soft label. Let $x$ be the position of the target point that we wish to classify. Compute  $D=\{d(X_i,x)\}_{i=1,...,M}$, the set of distances between each prototype and the target. Let $S'=(X_{(1)},Y_{(1)}), ...,(X_{(M)}, Y_{(M)}) $ be a reordering of the prototypes such that $d(X_{(1)},x)\leq ... \leq d(X_{(M)},x)$. Then the distance-weighted sum of the $k$-nearest prototype labels is $Y^*=\sum_{i=1}^k \frac{Y_{(i)}}{d(X_{(i)},x)}$ and $x$ is assigned to class $C^{SLaPkNN}(x) = \argmax_j Y^*_j$ where $Y^*_j$ is the $j^{th}$ element of $Y^*$.

\end{definition}

Distance-weighted kNN is the special case of SLaPkNN where each prototype is assigned a hard label.
\subsection{Probabilistic Prototypes and SLaPkNN with k=2 }
We first derive and analyze several methods of configuring soft-label prototypes to separate a space into $N$ partitions using $M$ points in the restricted case where prototypes must be probabilistic, and the number of considered neighbors ($k$) is set to two. 
\begin{theorem}{\textbf{(Learning Three Classes From Two Samples)}}
\label{thm:3_with_2}
Assume that two points are positioned 3 units apart in two-dimensional Euclidean space. Without loss of generality, suppose that point $x_1=(0,0)$ and point $x_2=(3,0)$ have probabilistic labels $y_1$ and $y_2$ respectively. We denote the $i^{th}$ element of each label by $y_{1,i}$ and $y_{2,i}$ for $i=1,2,3$. There exist $y_1$ and $y_2$ such that SLaPkNN with $k=2$ can separate three classes when fitted on $(x_1,y_1)$ and $(x_2,y_2)$. 
\end{theorem}
Assuming that we want symmetrical labels for the two prototypes (i.e. $y_{1,i}=y_{2,(3-i)}$), the resulting system of linear equations is quite simple.
\begin{align*}
    \begin{cases}
    \frac{2}{3}>y_{1,1}>\frac{1}{2}>y_{1,2}>\frac{1}{3}>y_{1,3}\geq 0\\
    4y_{1,2} = 1 + y_{1,1}\\
    5y_{1,2}=2 - y_{1,3}
\end{cases} \numberthis \label{eqn:3_inequalities_solution}
\end{align*}

Since we have a system of linear equations with one free variable, infinite solutions exist.
We set $y_{1,3}$ to zero in order to get a single solution and simplify the resulting label.
\begin{align*}
    y_{1,1}=y_{2,3}=\frac{3}{5}\\
    y_{1,2}=y_{2,2}=\frac{2}{5}\\
    y_{1,3}=y_{2,1}=\frac{0}{5}
\end{align*}
We visualize the results of fitting a SLaPkNN classifier with $k=2$ to a set of two points with these labels in Figure~\ref{fig:2points_3classes}.

\begin{corollary}
\label{cor:3_with_2}
Assume that the distance between two points (in two-dimensional Euclidean space) is $c$. Without loss of generality, suppose that point $x_1=(0,0)$ and point $x_2=(c,0)$ have probabilistic labels $y_1$ and $y_2$ respectively. We denote the $i^{th}$ element of each label by $y_{1i}$ and $y_{2i}$. There exist values of $y_1$ and $y_2$ such that SLaPkNN with $k=2$ can separate three classes when fitted on $(x_1,y_1)$ and $(x_2,y_2)$. 
\end{corollary}

\subsubsection{`Every Pair' Methods}
\begin{figure}[thbp]
\centering
\begin{subfigure}{.23\textwidth}
  \centering
  \includegraphics[width=\textwidth]{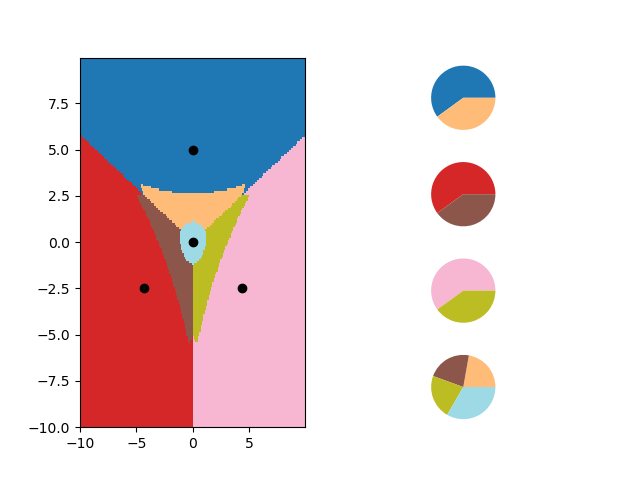}
  \caption{Seven classes using four soft-label prototypes}
  \label{fig:sub1c}
\end{subfigure}%
\hfill
\begin{subfigure}{.23\textwidth}
  \centering
  \includegraphics[width=\textwidth]{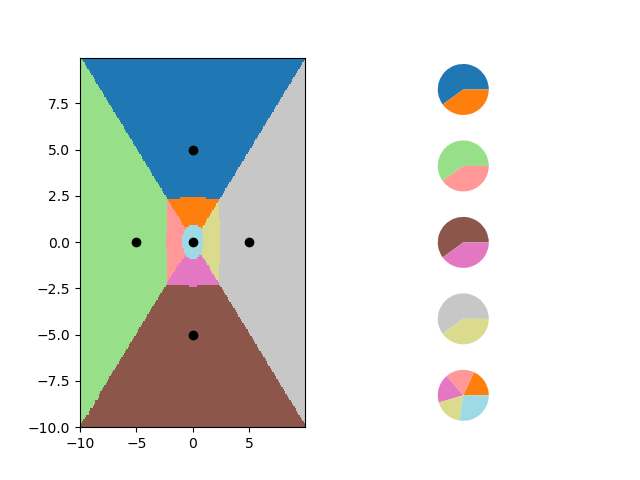}
  \caption{9 classes using 5 soft-label prototypes}
  \label{fig:sub2c}
\end{subfigure}
\caption{SLaPkNN can separate $2M-1$ classes using $M$ soft-label prototypes}
\label{fig:pairs_with_center}
\end{figure}

\begin{figure}[thbp]
\centering
\begin{subfigure}{.23\textwidth}
  \centering
  \includegraphics[width=\linewidth]{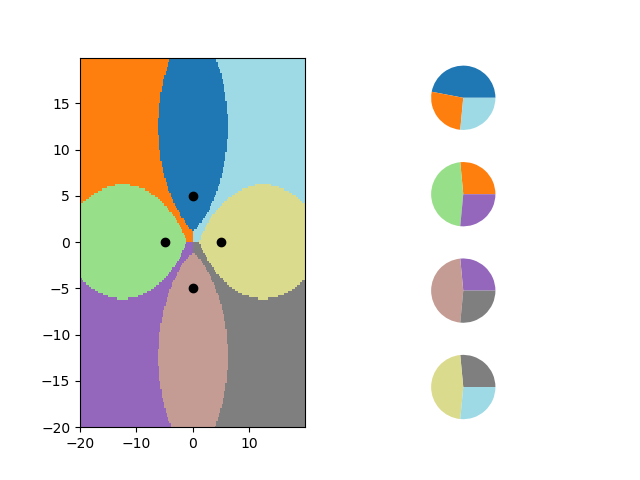}
  \caption{Eight classes using four soft-label prototypes}
  \label{fig:sub1a}
\end{subfigure}%
\hfill
\begin{subfigure}{.23\textwidth}
  \centering
  \includegraphics[width=\linewidth]{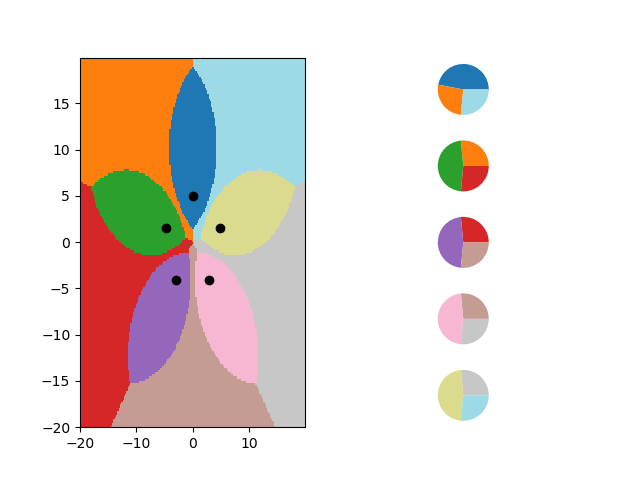}
  \caption{Ten classes using five soft-label prototypes}
  \label{fig:sub2a}
\end{subfigure}
\caption{SLaPkNN can separate $2M$ classes using $M$ soft-label prototypes}
\label{fig:every_other}
\end{figure}

We have shown that a third class can be induced between a pair of points. We now focus on the case where we consider more than two points.  We first show that it is possible for a single point to simultaneously belong to multiple pairs, each creating an additional class. 
\begin{theorem} \label{1point_many_pairs}
Suppose we have $M$ soft-label prototypes $(x_0,y_0), (x_1,y_1), ... (x_{M-1},y_{M-1})$ with the $x_i$ arranged such that each pair of the form $\{(x_0,x_i)| i=1,..,M-1\}$ is unique and the other terms are all equidistant from $x_0$. There exist values of $y_0, y_1, ..., y_{M-1}$ such that SLaPkNN with $k=2$ can can partition the space into $2M-1$ classes.
\end{theorem}
One way to select such labels is to use the same labels for each pair as in Theorem~\ref{thm:3_with_2}, This results in $y_1,...,y_{M-1}$ each having a label distribution containing two non-zero values: $\frac{3}{5}$ (associated with its main class) and $\frac{2}{5}$ (associated with the class created between itself and $x_0$). Meanwhile, $y_0$ contains one element with value $\frac{3}{5}$ (associated with its own class) and M-1 elements with value $\frac{2}{5}$ (each associated with a unique class created between $x_0$ and each one of the other points). To get probabilistic labels, we can normalize $y_0$ to instead have values $\frac{3}{2M+1}$ and $\frac{2}{2M+1}$. The resulting decision landscape is visualized in Figure~\ref{fig:pairs_with_center}. The local decision boundaries in the neighbourhood between $(x_0,y_0)$ and any one of the surrounding prototypes then takes the following form.
\begin{align}
    \text{Predicted Class}=\begin{cases} 
    a \text{ if } d<\frac{5p}{4M+7}\\
    b \text{ if } d>\frac{10p}{2M+11}\\
    c \text{ if } \frac{5p}{4M+7}<d<\frac{10p}{2M+11}
    %c \text{ if } \frac{1}{(2M+1)d}<\frac{2}{5(p-d)}, \frac{1}{5(p-d)}<\frac{2}{(2M+1)d}
    \end{cases}
\end{align}
Examining the asymptotic behavior as the total number of classes increases, we notice a potentially undesirable property of this configuration.
% \begin{align}
%     \lim_{M\rightarrow\infty}
%     \begin{cases} 
%     a \text{ if } d<\frac{5p}{4M+7} \rightarrow d<0 \\
%     b \text{ if } d>\frac{10p}{2M+11} \rightarrow d>0 \\
%     c \text{ if } \frac{5p}{4M+7}<d<\frac{10p}{2M+11} \rightarrow 0<d<0
%     \end{cases}
% \end{align}
Increases in $M$ `dilute' classes $a$ and $c$, which results in them shrinking towards $x_0$. In the limit, only class $b$ remains.
It is possible to find a configuration of probabilistic labels that results in asymptotically stable classes but this would require either lifting our previous restriction that the third label value be zero when separating three classes with two points, or changing the underlying geometrical arrangement of our prototypes.
\begin{proposition}\label{prop:every_pair}
Suppose $M$ soft-label prototypes are arranged as the vertices of an $M$-sided regular polygon. There exist soft labels $(Y_1,...,Y_M)$ such that fitting SLaPkNN with $k=2$ will divide the space into $2M$ classes.
\end{proposition}
In this configuration, it is possible to decouple the system from the number of pairs that each prototype participates in. It can then be shown that the local decision boundaries, in the neighbourhood of any pair of adjacent prototypes, do not depend on M. 
\begin{align}
    \text{Predicted Class}=\begin{cases} 
    a \text{ if } d<\frac{p}{3}\\
    b \text{ if } d>\frac{2p}{3}\\
    c \text{ if } \frac{p}{3}<d<\frac{2p}{3}
    %c \text{ if } \frac{1}{(2M+1)d}<\frac{2}{5(p-d)}, \frac{1}{5(p-d)}<\frac{2}{(2M+1)d}
    \end{cases}
\end{align}
We visualize the resulting decision landscape in Figure~\ref{fig:every_other}.
By combining these two results, we can produce configurations that further increase the number of classes that can be separated by $M$ soft-label prototypes.
\begin{theorem}\label{cor:pairs_with_center}
Suppose $M$ soft-label prototypes are arranged as the vertices and center of an $(M-1)$-sided regular polygon. There exist soft labels $(Y_1,...,Y_M)$ such that fitting SLaPkNN with $k=M$ will divide the space into $3M-2$ classes.
\end{theorem}

\begin{figure}[thbp]
\centering
\begin{subfigure}{.23\textwidth}
  \centering
  \includegraphics[width=\linewidth]{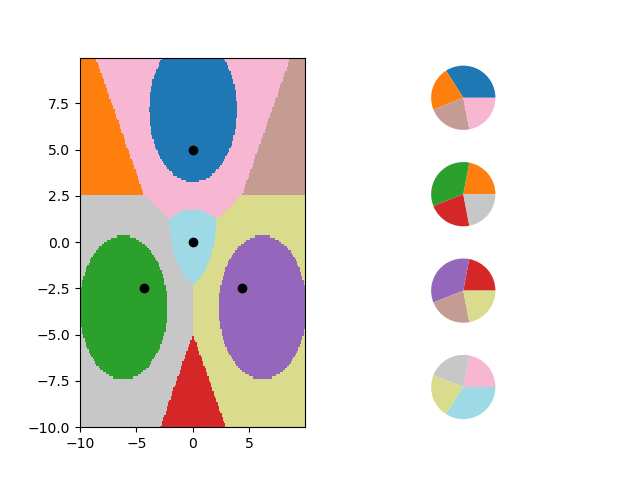}
  \caption{Ten classes using four soft-label prototypes}
  \label{fig:sub1b}
\end{subfigure}%
\hfill
\begin{subfigure}{.23\textwidth}
  \centering
  \includegraphics[width=\linewidth]{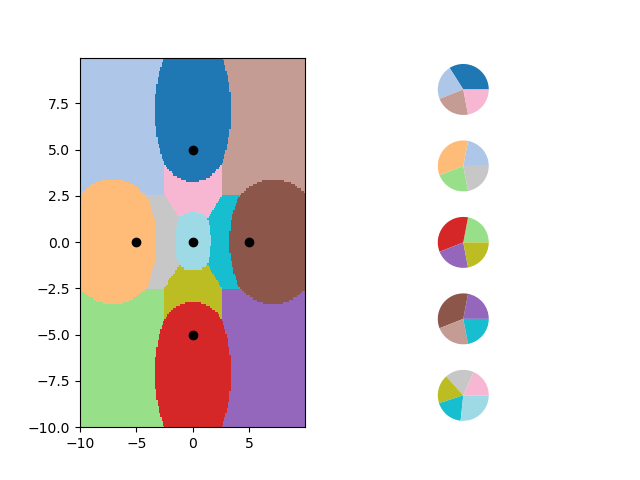}
  \caption{Thirteen classes using five soft-label prototypes}
  \label{fig:sub2b}
\end{subfigure}
\caption{SLaPkNN can separate $3M-2$ classes using $M$ soft-label prototypes}
\label{fig:test}
\end{figure}

% \begin{definition}
% [NEED TO DEFINE True centroid of a class vs the apparent centroid of a class.]
% \end{definition}
% \begin{proposition}
% Suppose $M$ soft-label prototypes are arranged such that the centroid of every subset of prototypes is unique. There exist soft labels $(Y_1,...,Y_M)$ such that fitting SLaPkNN with $k=M$ will partition the space into $2^M-1$ partitions.
% \end{proposition}
% \begin{proof}

% We proceed in reverse from our previous strategy. 

% [WRITE THIS ONE UP]

% \end{proof}
% \begin{figure}[h]
% \centering
% \begin{subfigure}{.23\textwidth}
%   \centering
%   \includegraphics[width=\linewidth]{images/4_points_10_classes.png}
%   \caption{Fifteen classes using four soft-label prototypes}
%   \label{fig:sub1}
% \end{subfigure}%
% \hfill
% \begin{subfigure}{.23\textwidth}
%   \centering
%   \includegraphics[width=\linewidth]{images/5_points_13_classes.png}
%   \caption{Thirty-one classes using five soft-label prototypes}
%   \label{fig:sub2}
% \end{subfigure}
% \caption{[PLACEHOLDER IMAGES] SLaPkNN can separate $2^M-1$ classes using $M$ soft-label prototypes}
% \label{fig:test}
% \end{figure}

\subsubsection{Interval Partitioning Methods}
We now aim to show that two points can even induce multiple classes between them.

\begin{figure}[h!]
    \centering
    \includegraphics[width=0.48\textwidth]{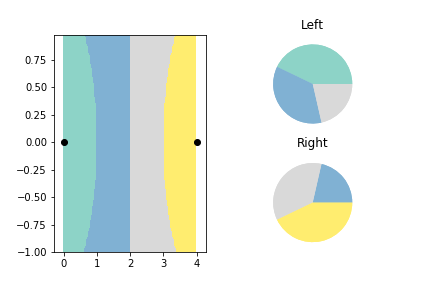}
    \caption{A SLaPkNN classifier is fitted on two points and used to partition the space into four classes. The probabilistic soft labels of each point are illustrated by the pie charts.}
    \label{fig:2points_4classes}
\end{figure}

\begin{lemma}
\label{thm:4_with_2}
Assume that two points are positioned 4 units apart in two-dimensional Euclidean space. Without loss of generality, suppose that point $x_1=(0,0)$ and point $x_2=(4,0)$ have probabilistic labels $y_1$ and $y_2$ respectively. We denote the $i^{th}$ element of each label by $y_{1,i}$ and $y_{2,i}$. There exist values of $y_1$ and $y_2$ such that SLaPkNN with $k=2$ can separate four classes when fitted on $x_1$ and $x_2$.  
\end{lemma}
We can again find a solution to this system that has symmetrical labels and also sets an element of the label to zero.
\begin{align*}
    y_{1,1}=y_{2,4}=\frac{6}{14}\\
    y_{1,2}=y_{2,3}=\frac{5}{14}\\
    y_{1,3}=y_{2,2}=\frac{3}{14}\\
    y_{1,4}=y_{2,1}=\frac{0}{14}\\
\end{align*}
We visualize the results of fitting a SLaPkNN classifier with $k=2$ to a set of two points with these labels in Figure~\ref{fig:2points_4classes}.

We can further extend this line of reasoning to produce the main theorem of this paper.
\begin{theorem}{\textbf{(Main Theorem)}}
SLaPkNN with $k=2$ can separate $n \in [1,\infty)$ classes using two soft-label prototypes. 
\label{thm:n_with_2}
\end{theorem}
This unexpected result is crucial for enabling extreme LO-shot learning as it shows that in some cases we can completely disconnect the number of classes from the number of prototypes required to separate them. The full proof can be found in the supplemental materials, but we provide a system of equations describing soft labels that result in two soft-label prototypes separating  $n \in [1,\infty)$ classes. 
\begin{align*}
    y_{1,i}&= y_{2,(n-i)}=\frac{\sum_{j=i}^{n-1}j}{\sum_{j=1}^{n-1}j^2} = \frac{n(n-1)-i(i-1)}{2\sum_{j=1}^{n-1}j^2}, \\ i&=1,2,...,n
\end{align*}

% \begin{corollary}
% SLaPkNN can separate $mn \in [0,\infty)$ classes using $m$ soft-label prototypes
% \end{corollary}
\subsection{Other Results}
The diverse decision landscapes we have already seen were generated using probabilistic labels and $k=2$. Using unrestricted soft labels and modifying $k$ enables us to explore a much larger space of decision landscapes. Due to space constraints we present only a few of the results from our experiments with modifying these parameters, but more are included in the accompanying supplemental material. One unexpected result is that SLaPkNN can generate any number of concentric elliptical decision bounds using a fixed number of prototypes as seen in Figures~\ref{fig:ellipses} and~\ref{fig:2points}. In Figure~\ref{fig:8_15_kprogression}, we show that variations in $k$ can cause large changes in the decision landscape, even producing non-contiguous classes.
% \begin{figure}
%     \centering
%     \includegraphics[width=0.48\textwidth]{images/3points_8rings.png}
%     \caption{SLaPkNN can create any number of ellipses using 3 soft-label prototypes. Each pie chart represents the soft label of one prototype.}
%     \label{fig:my_label}
% \end{figure}

\subsection{Robustness}
In order to understand the robustness of LO-shot learning landscapes to noise, we aim to analyze which regions are at highest risk of changing their assigned class if the prototype positions or labels are perturbed. We use the difference between the largest predicted label value and second-largest predicted label value as a measure of confidence in our prediction for a given point. The risk of a given point being re-classified is inversely proportional to the confidence. Since our aim is to inspect the entire landscape at once, rather than individual points, we visualize risk as a gradient from black (high confidence/low risk) to white (low confidence/high risk) over the space. We find that due to the inverse distance weighting mechanism, there are often extremely high confidence regions directly surrounding the prototypes, resulting in a very long-tailed, right-skewed distribution of confidence values. As a result, we need to either clip values or convert them to a logarithmic scale in order to be able to properly visualize them. We generally find that clipping is helpful for understanding intra-class changes in risk, while the log scale is more suited for visualizing inter-class changes in risk that occur at decision boundaries.

\begin{figure*}
    \centering
\begin{subfigure}[t]{0.49\textwidth}%[tp!]
    %\centering
    \includegraphics[width=\textwidth]{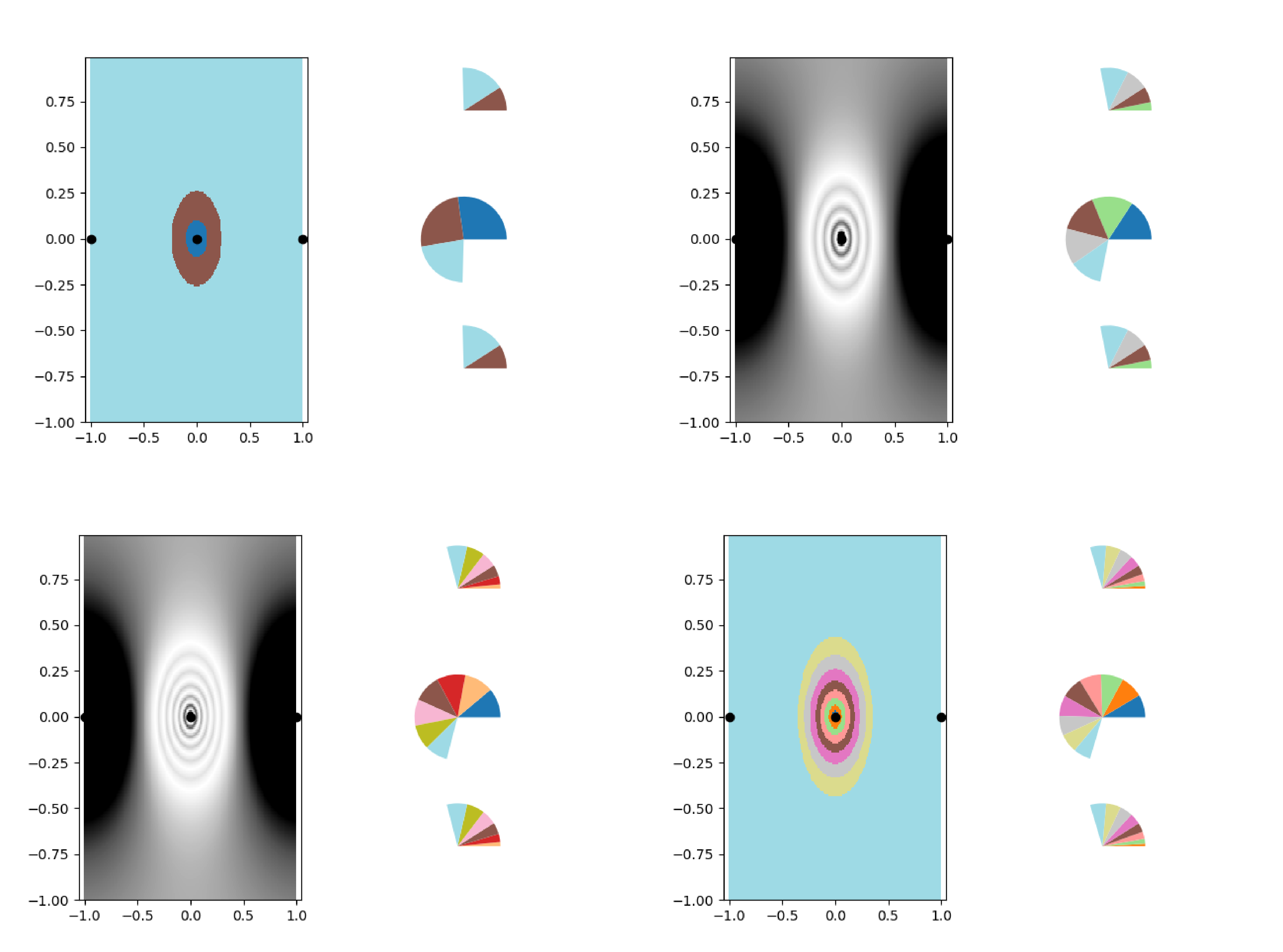}
    \caption{SLaPkNN with $k=3$ is shown separating $N=3, 5, 7, 9$ classes (left to right, top to bottom) with elliptical decision boundaries after being fitted on 3 prototypes. The pie charts represent unrestricted labels.}
    \label{fig:ellipses}
\end{subfigure}
\hfill
\begin{subfigure}[t]{0.49\textwidth}%[p!]
    %\centering
    \includegraphics[width=\textwidth]{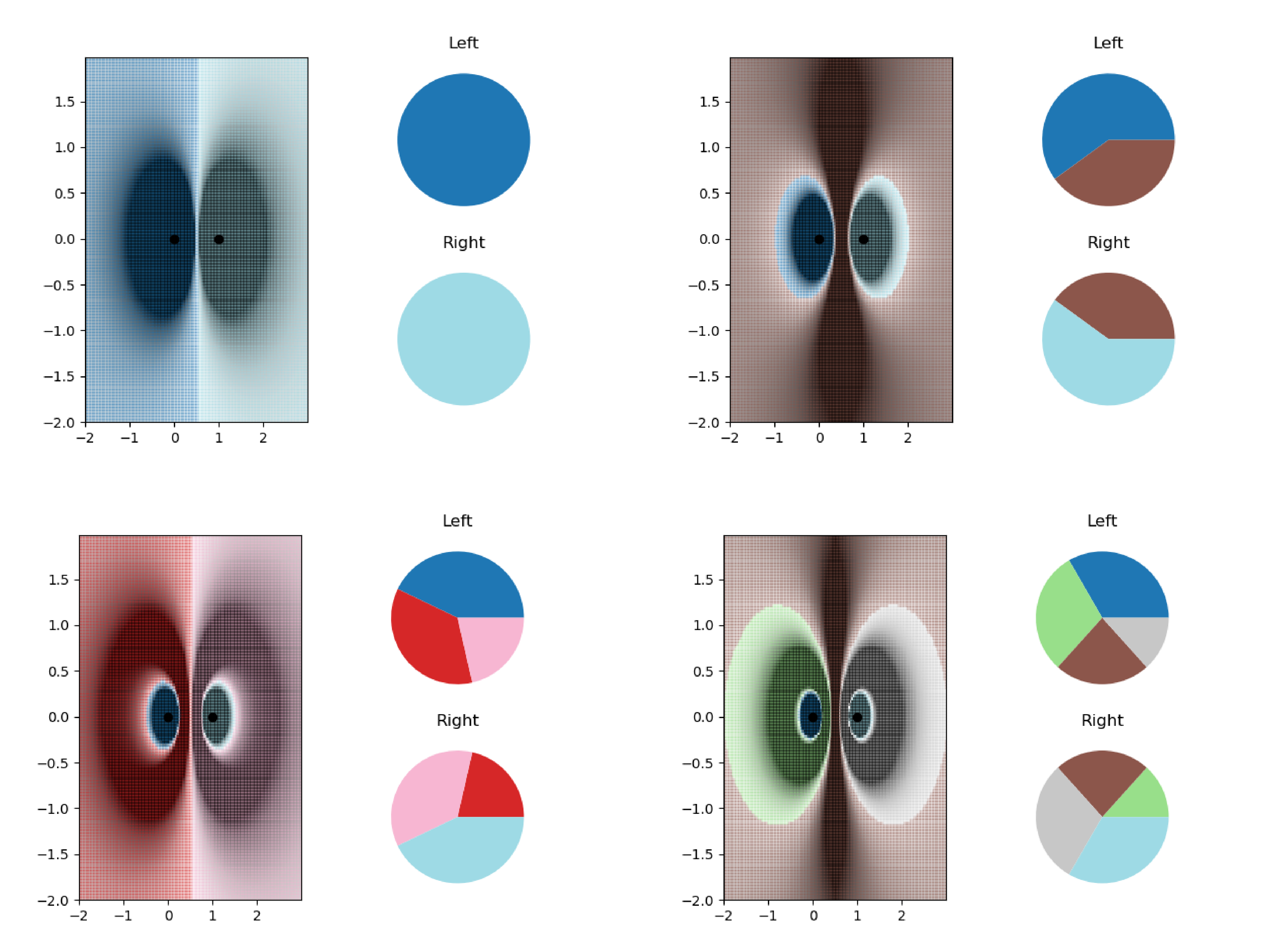}
    \caption{SLaPkNN with $k=2$ is shown separating $N=2, 3, 4, 5$ classes (left to right, top to bottom) after being fitted on 2 prototypes. The pie charts represent probabilistic labels.}
    \label{fig:2points}
\end{subfigure}
\begin{subfigure}[b]{0.49\textwidth}%[p!]
    %\centering
    \includegraphics[width=\textwidth]{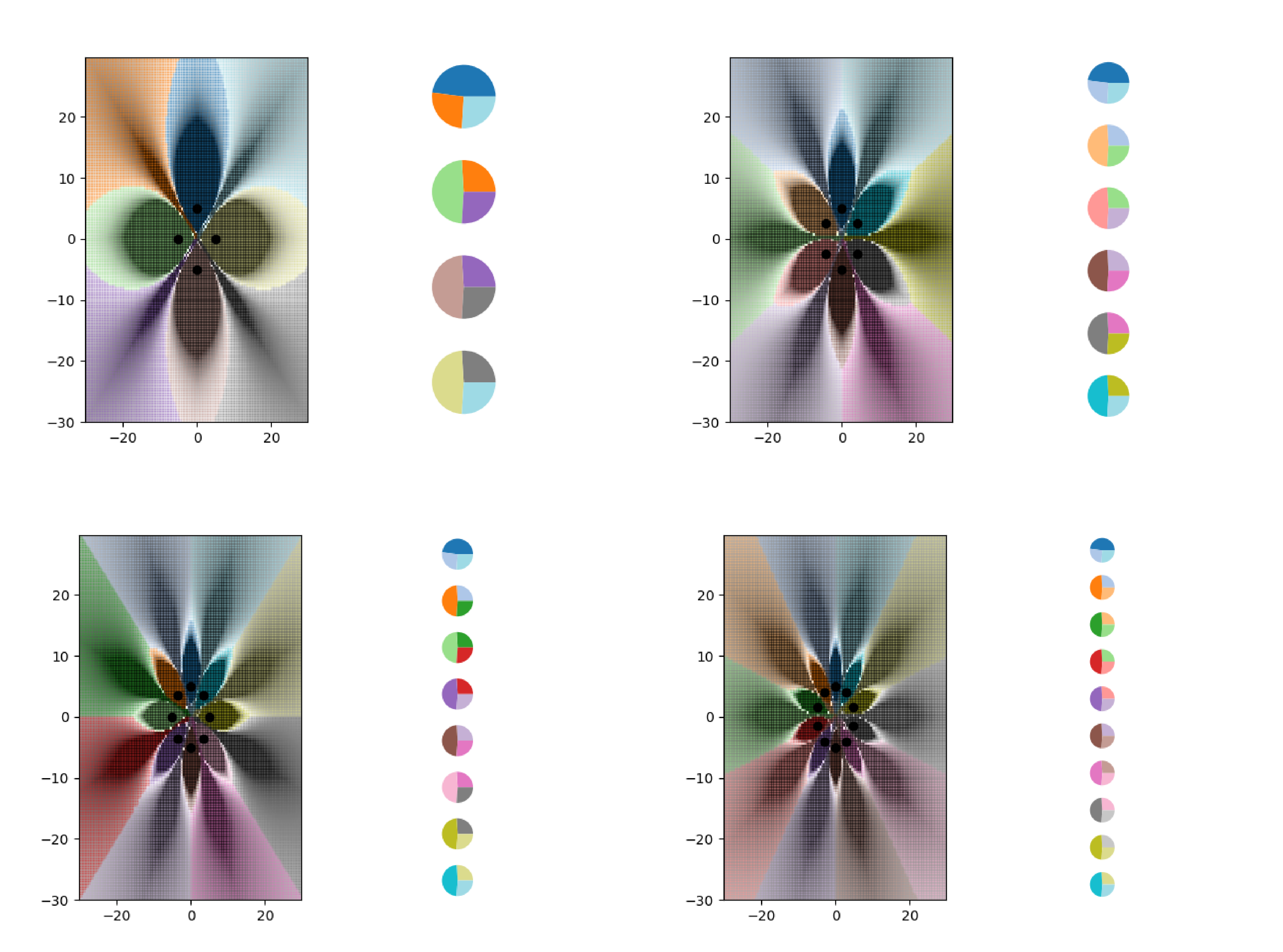}
    \caption{SLaPkNN with $k=2$ is shown separating $2M$ classes after being fitted on $M=4,6,8,10$ prototypes (left to right, top to bottom).  The pie charts represent probabilistic labels.}
    \label{fig:pairs_Mprogression}
\end{subfigure}
\hfill
\begin{subfigure}[b]{0.49\textwidth}%[p!]
    %\centering
    \includegraphics[width=\textwidth]{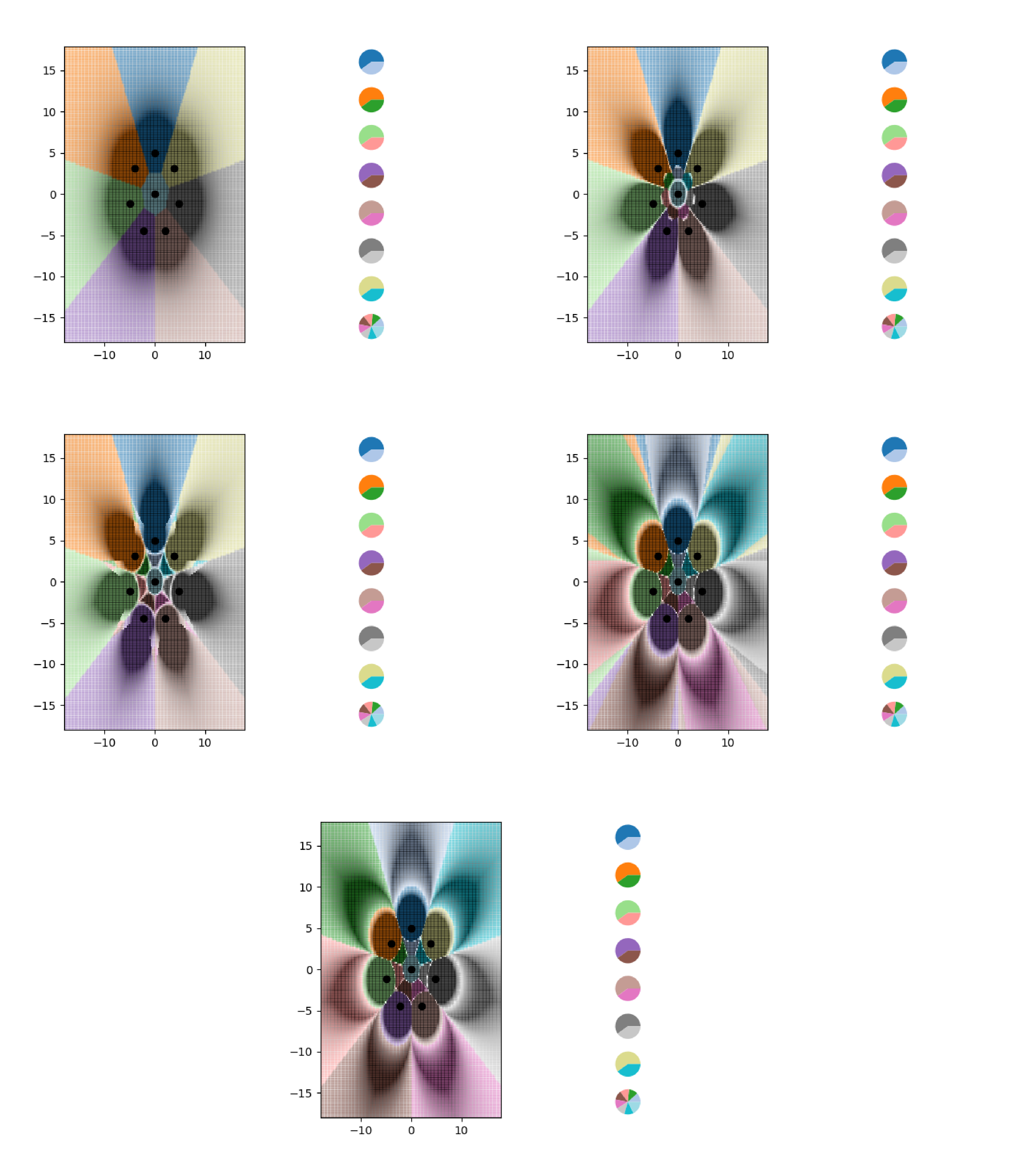}
    \caption{SLaPkNN with $k=1, 2, 3, 4, 5$ (left to right, top to bottom) is shown separating 15 classes after being fitted on 8 prototypes. The pie charts represent unrestricted labels.}
    \label{fig:8_15_kprogression}
\end{subfigure}
\caption{Various LO-shot learning decision landscapes and risk gradients are presented. Each color represents a different class. Gray-scale is used to visualize the risk gradient, with darker shadows corresponding to lower risk. In (a), the two colorful charts show decision landscapes and the two gray-scale charts show risk landscapes. In (b)-(d), the risk gradient is laid directly over the decision landscape. The pie charts represent the soft labels of each prototype.}
\label{fig:landscapes_and_grads}
\end{figure*}

Figure~\ref{fig:landscapes_and_grads} visualizes the risk gradient for many of the decision landscapes derived above. Overall, we find that LO-shot learning landscapes have lower risk in regions that are distant from decision boundaries and lowest risk near the prototypes themselves. Changes in $k$ can have a large effect on the inter-class risk behaviour which can be seen as changes in the decision boundaries. However, they tend to have a smaller effect on intra-class behavior in regions close to the prototypes. The most noticeable changes occur when increasing $k$ from 1 to 2, as this causes a large effect on the decision landscape itself, and from 2 to 3, which tends to not affect the decision landscape but causes a sharp increase in risk at decision boundaries. Meanwhile, increasing the number of classes ($M$) can in some cases cause large changes in the intra-class risk behaviour, but we can design prototype configurations (such as the ellipse generating configuration) that prevent this behaviour. In general, it appears that by tuning the prototype positions and labels we can simultaneously elicit desired properties both in the decision landscape and risk gradient. This suggests that LO-shot learning is not only feasible, but can also be made at least partially robust to noisy training data.

\subsection{Case Study: Prototype Generation for Circles}
We consider a simple example that captures the difference between hard and soft-label prototype generation. Assume that our data consists of concentric circles, where each circle is associated with a different class. We wish to select or generate the minimal number of prototypes such that the associated kNN classification rule will separate each circle as a different class. We tested several hard-label prototype methods and found their performance to be poor on this simulated data. Many of the methods produced prototypes that did not achieve separation between classes which prevents us from performing a fair comparison between hard and soft-label prototypes.  In order to allow such a comparison, we analytically derive a near-optimal hard-label prototype configuration for fitting a 1NN classifier to the circle data. 

\begin{figure}[t!]
    \centering
    \includegraphics[width=0.45\textwidth]{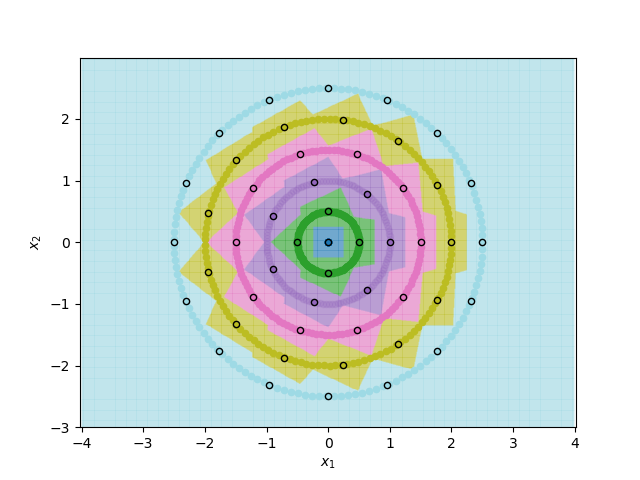}
    \caption{Decision boundaries of a vanilla 1NN classifier fitted on the minimum number of prototypes required to perfectly separate circle classes. From inner to outer, the circles have 1, 4, 7, 10, 13, and 16 prototypes.}
    \label{fig:knn_circles}
\end{figure}

\begin{figure}[t!]
    \centering
    \includegraphics[width=0.45\textwidth]{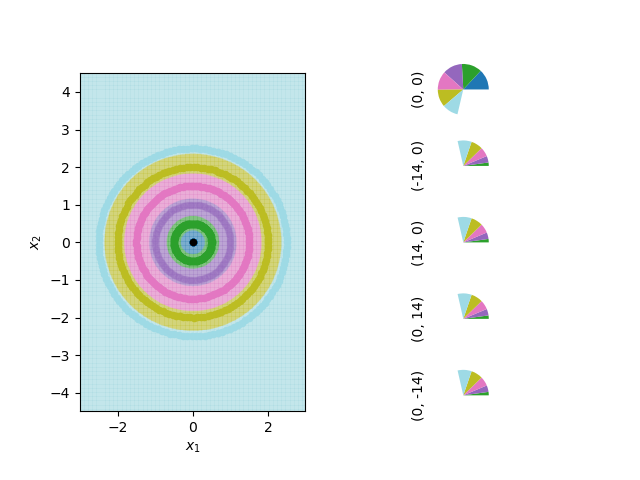}
    \caption{SLaPkNN can separate 6 circles using 5 soft-label prototypes. Each pie chart represents the soft label of one prototype, and is labeled with its location. 4 of the prototypes are located outside of the visible range of the chart.}
    \label{fig:5_circles_5_points}
\end{figure}

\begin{theorem}
\label{thm:knn_circles_upperbound}
Suppose we have $N$ concentric circles with radius $r_t=t*c$ for the $t^{th}$ circle. An upper bound for the minimum number of hard-label prototypes required for 1NN to produce decision boundaries that perfectly separate all circles, is $\sum_{t=1}^{N} \frac{\pi}{\cos ^{-1}\left(1-\frac{1}{2 t^{2}}\right)}$; with the $t^{th}$ circle requiring $\frac{\pi}{\cos ^{-1}\left(1-\frac{1}{2 t^{2}}\right)}$ prototypes. 
\end{theorem}
 The proof can be found in the appendix. We can use the approximation $cos^{-1}(1-y)\approx\sqrt{2y}$ to get that $\frac{\pi}{cos^{-1}(1-\frac{1}{2t^2})}\approx\frac{\pi}{(\frac{1}{t})}=t\pi$. Thus the upper bound for the minimum number of prototypes required is approximately $\sum_{t=1}^{N}t*\pi = \frac{N(N+1)\pi}{2}$. Figure~\ref{fig:knn_circles} visualizes the decision boundaries of a regular kNN that perfectly separates six concentric circles given $\ceil{t\pi}$ prototypes for the $t^{th}$ circle. It is possible that the number of prototypes may be \textit{slightly} reduced by carefully adjusting the prototype locations on adjacent circles to maximize the minimal distance between the prototype midpoints of one circle and the prototypes of neighboring circles. 
 However, SLaPkNN requires only a constant number of prototypes to generate concentric ellipses. In Figure~\ref{fig:5_circles_5_points}, SLaPkNN fitted on five soft-label prototypes is shown separating the six circles. The decision boundaries created by SLaPkNN match the underlying geometry of the data much more accurately than those created by 1NN. This is because 1NN can only create piecewise-linear decision boundaries.

In this case study, enabling soft labels reduced the minimal number of prototypes required to perfectly separate $N$ classes from $\mathcal{O}(N^2)$ down to $\mathcal{O}(1)$. We note that the number of required hard-label prototypes may be reduced by carefully tuning $k$ as well as the neighbor weighting mechanism (e.g. uniform or distance-based). However, even in the best case, the number of required hard-label prototypes is at the very least linear in the number of classes.

% \paragraph{Palmer Penguins}
% [INSERT PENGUINS RESULTS]
\section{Conclusion}
We have presented a number of results that we believe can be used to create powerful and efficient dataset condensation and prototype generation techniques. More generally, our contributions lay the theoretical foundations necessary to establish `less than one'-shot learning as a viable new direction in machine learning research. We have shown that even a simple classifier like SLaPkNN can perform LO-shot learning, and we have proposed a way to analyze the robustness of the decision landscapes produced in this setting. We believe that creating a soft-label prototype generation algorithm that specifically optimizes prototypes for LO-shot learning is an important next step in exploring this area. Such an algorithm will also be helpful for empirically analyzing LO-shot learning in high-dimensional spaces where manually designing soft-label prototypes is not feasible. 

Additionally, we are working on showing that LO-shot learning is compatible with a large variety of machine learning models. Improving prototype design is critical for speeding up instance-based, or lazy learning, algorithms like kNN by reducing the size of their training sets. However, eager learning models like deep neural networks would benefit more from the ability to learn directly from a small number of real samples to enable their usage in settings where little training data is available. This remains a major open challenge in LO-shot learning.

\bibliography{references}

\end{document}